\documentclass[10pt,twocolumn,letterpaper]{article}

\usepackage[pagenumbers]{cvpr} %
\usepackage{float}
\usepackage{booktabs}
\usepackage{wrapfig}
\usepackage{listings}
\usepackage{algorithm}
\usepackage{algpseudocode}
\usepackage{amsmath}
\usepackage{svg}
\usepackage{graphicx}
\usepackage{subcaption}
\usepackage{multirow}

\makeatletter
\@namedef{tabular*}#1{\footnotesize%
 \setlength\dimen@{#1}%
   \edef\@halignto{to\the\dimen@}\@tabular}

\def\Hline{%
  \noalign{\ifnum0=`}\fi\hrule \@height .43pt \futurelet%
   \@tempa\@xhline}
\makeatother

\definecolor{cvprblue}{rgb}{0.21,0.49,0.74}
\usepackage[pagebackref,breaklinks,colorlinks,citecolor=cvprblue]{hyperref}

\title{Open-vocabulary 3D scene perception in industrial environments}

\author{
Keno Moenck\textsuperscript{\rm 1,}\thanks{Corresponding author: \href{mailto:keno.moenck@tuhh.de}{keno.moenck@tuhh.de}} \quad Adrian Philip Florea\textsuperscript{\rm 1} \quad Julian Koch\textsuperscript{\rm 1} \quad Thorsten Sch{\"u}ppstuhl\textsuperscript{\rm 1} \vspace{5pt}\\
{\normalsize \textsuperscript{\rm 1}{Hamburg University of Technology, Institute of Aircraft Production Technology}} \vspace{5pt}
}

\begin{document}
\maketitle

\begin{abstract}
Autonomous vision applications in production, intralogistics, or manufacturing environments require perception capabilities beyond a small, fixed set of classes.
Recent open-vocabulary methods, leveraging 2D Vision-Language Foundation Models (VLFMs), target this task but often rely on class-agnostic segmentation models pre-trained on non-industrial datasets (e.g., household scenes).
In this work, we first demonstrate that such models fail to generalize, performing poorly on common industrial objects. Therefore, we propose a training-free, open-vocabulary 3D perception pipeline that overcomes this limitation.
Instead of using a pre-trained model to generate instance proposals, our method simply generates masks by merging pre-computed superpoints based on their semantic features.
Following, we evaluate the domain-adapted VLFM "IndustrialCLIP" on a representative 3D industrial workshop scene for open-vocabulary querying. Our qualitative results demonstrate successful segmentation of industrial objects.
\end{abstract}

\section{Introduction}\label{sec:introduction}
Visual perception\footnote{On purpose, we summarize tasks, such as semantic, instance, or panoptic segmentation as well as object recognition, under visual perception -- the interpretation of visual sensory information.} in an environment's three-dimensional representation enables various kinds of autonomous online or offline applications. In offline settings, the representation can be referred to as a geometric digital twin \cite{moenckGeometricDigitalTwins2024a}, which can serve, for example, progress monitoring during construction or the planning of retrofits. In contrast, in online settings, the representation may only partially cover the scene and is updated during exploration, while allowing an autonomous agent to plan actions.

\begin{figure}
    \centering
    \includegraphics[width=0.75\linewidth]{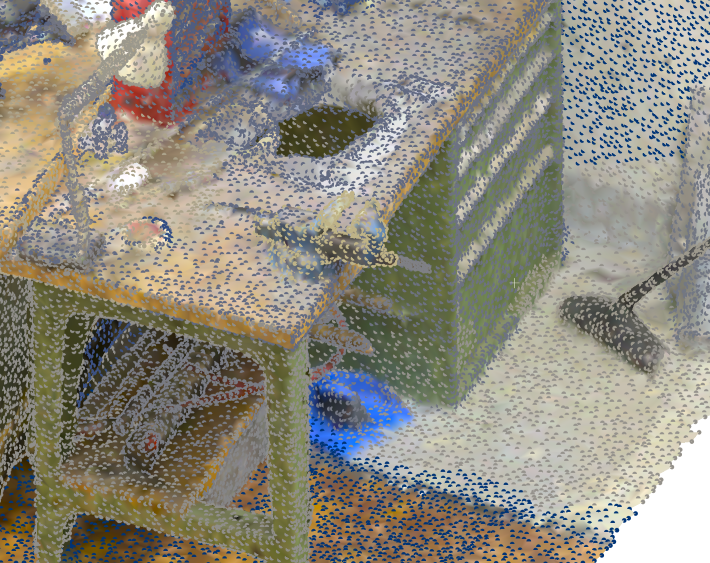}
    \includegraphics[width=0.75\linewidth]{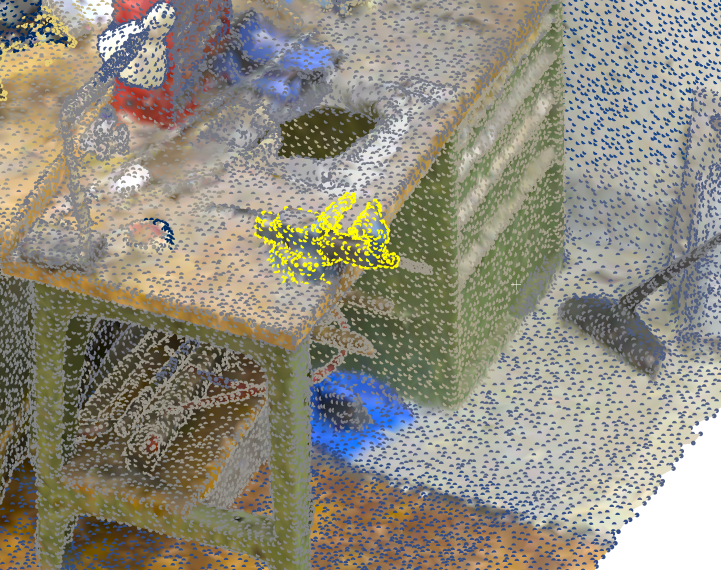}
    \caption{Prompting "vise" using features from CLIP and IndustrialCLIP (yellow and blue-colored points correspond to high and low semantic similarity scores, respectively).}
    \label{fig:qualitative_clip_iclip}
\end{figure}

3D machine learning-based perception in unique domains faces a primary bottleneck: data scarcity, as most public datasets lack domain-specific relevance.
While there are multiple 3D indoor and outdoor datasets for training perception models, for Building Information Modeling (BIM) \cite{rozenberszkiLanguageGroundedIndoor3D2022} and autonomous driving \cite{liaoKITTI360NovelDataset2023}, there are no comprehensive real-world publicly available datasets for training perception models in industrial environments, which represent production, intralogistics, or manufacturing environments.
Traditionally, one could address this issue via synthetic data engineering~\cite{wuSim2realTransferLearning2023,sommerAutomatedGenerationDigital2023,abouakarSORDIaiLargescaleSynthetic2025,noichlAutomatedMethodsCreating2025} or data-efficient strategies, such as one-shot or few-shot learning~\cite{zhaoFewshot3DPoint2021}. However, for the former, the data engineering effort is seldom economical when targeting new object classes or higher robustness, while powerful foundational models that enable the latter are still under research.

Recent advances in computer vision have been dominated by the self-supervised pre-training of foundation models, which serve as bases for a range of diverse downstream tasks. The success of Large Language Models (LLMs) and Vision-Language Foundation Models (VLFMs) is directly attributed to self-supervised pre-training on web-scale data. However, web-scale data is limited to the text and 2D image/video modalities.
A solution to exploit 2D VLFM in the 3D domain is "lifting" them by using the 3D scan's posed 2D images \cite{pengOpenScene3DScene2023b,takmazOpenMask3DOpenvocabulary3D2023b,nguyenOpen3DISOpenVocabulary3D2024a,jungDetailsMatterIndoor2025}.
Further, VLFMs enable perception capabilities that extend beyond a fixed set of classes, as learned in fully supervised settings. In this open-vocabulary setting, the set of "classes" is not limited. This means that we can prompt using natural language, including not only simple words, but also properties or affordances, such as "red pliers" or "clean workbench".

A common paradigm in open-vocabulary 3D perception involves a multi-stage pipeline. The approaches typically begin by partitioning the 3D scene using pre-trained class-agnostic instance segmentation modules. Then, a best view selection algorithm identifies the most informative 2D perspectives for each segmented instance. A 2D VLFM subsequently processes these selected views. Finally, the semantic features or similarity scores derived from these views are merged or aggregated to assign a robust open-vocabulary quantity for each object.

In this work, we follow the idea of "lifting" 2D VLFM into the 3D domain and evaluate the performance of an adapted 2D VLFM for industrial settings, namely IndustrialCLIP \cite{moenckIndustrialLanguageImageDataset2024b}, as depicted in Fig.~\ref{fig:qualitative_clip_iclip}.
However, during preliminary experiments, we found that a supervised pre-trained class-agnostic instance segmentation module does not perform well on a scene from the industrial context, which is why we propose a new method of generating mask proposals based on feature-based merging of superpoints (pre-computed cluster of points). Our contributions are as follows:
\begin{itemize}
    \item Demonstration of training-free, open-vocabulary 3D perception (instance and semantic segmentation) in a representative industrial scene;
    \item Qualitative evaluation of the performance and limitations of using IndustrialCLIP for 3D perception;
    \item Evaluation of the replacement of overfitted class-agnostic instance segmentation models with superpoints and a feature-based merging strategy.
\end{itemize}

The remainder of this work is structured as follows: First, in Sec.~\ref{sec:fundamentals} and \ref{sec:related_works}, we provide an overview of fundamentals and related works.
In Sec.~\ref{sec:preliminary}, we present the results of our preliminary experiments.
In Sec.~\ref{sec:method}, we present the overall method that led to our experiments, and the results of which we present in Sec.~\ref{sec:experiments}.
Finally, we conclude and discuss in Sec.~\ref{sec:discussion}.

\section{Fundamentals}\label{sec:fundamentals}

\textit{Contrastive language-image pre-training} is a self-supervised pre-training method, in which language-image samples are embedded through a text and an image encoder, respectively. During training, embeddings are typically aligned using a cosine similarity loss. On a larger scale, it was initially prominently demonstrated by CLIP \cite{radfordLearningTransferableVisual2021a} and ALIGN \cite{jiaScalingVisualVisionLanguage2021}. CLIP was trained on 400M samples, learning rich, generalizable representations from unorganized text-image samples from the web, establishing the entire model or even only the vision or text encoder as frequently used.

\textit{IndustrialCLIP} refers to an adapted CLIP model for industrial settings by transfer learning on the Industrial Language-Image Dataset (ILID) \cite{moenckIndustrialLanguageImageDataset2024b}, generated by web-crawling online catalogs for industrial objects. Parameter-efficient Fine-tuning (PEFT) was performed via prompt learning and residual-based adapter-style image embedding fine-tuning.
IndustrialCLIP significantly outperforms the baseline zero-shot CLIP in industrial applications, such as correctly identifying components or language-guided 2D segmentation \cite{moenckIndustrialLanguageImageDataset2024b}.

\textit{Segment Anything} introduced a foundation model for promptable, class-agnostic 2D segmentation \cite{kirillovSegmentAnything2023a}. Using an image encoder, a prompt encoder, and a mask decoder, it partitions images into semantically meaningful whole/part/subpart masks per prompt.
The resulting model's outstanding generalization capabilities are described as the "GPT-3 moment" of computer vision.
The Segment Anything Model (SAM) learned a general concept of objectness that generalizes well across domains.

\section{Related works}\label{sec:related_works}

\subsection{Open-vocabulary 3D scene understanding}\label{sec:related_works_open_vocab}

Recent work has brought about a transition from closed-vocabulary to handling open-vocabulary queries with no restrictions on the input text.
In the following, we outline the core contributions in open-vocabulary 3D perception in historical order. Besides the mentioned methods and models, multiple further derivatives exist, which, however, share the same outlined limitations, making them unsuitable for the purposes of this work.

OpenScene \cite{pengOpenScene3DScene2023b} is the first approach in open-vocabulary 3D segmentation based on a 3D encoder that outputs per-point features aligned with CLIP embeddings. During training, image features are extracted pixel-wise for each projected 3D point across the posed 2D images, fused, and used as targets for the 3D encoder.
During inference, the 3D encoder outputs per-point CLIP embedding space-aligned features, which are contrasted with text embeddings, yielding per-point similarity scores per text query.
OpenScene requires a training process on data (from the respective domain) to probably tune the 3D encoder.

OpenMask3D \cite{takmazOpenMask3DOpenvocabulary3D2023b} follows the idea of first segmenting a scene into class-agnostic proposals and subsequently projecting these into posed 2D images and extracting CLIP-aligned features per object proposal.
Mask3D \cite{schultMask3DMaskTransformer2023a} serves as the class-agnostic mask proposal model. It is pre-trained on the same dataset (ScanNet200 \cite{rozenberszkiLanguageGroundedIndoor3D2022}) as used in the evaluation of OpenMask3D. Since OpenMask3D does not optimize the initially generated mask proposals, performance is directly tied to Mask3D's capabilities.

Open3DIS \cite{nguyenOpen3DISOpenVocabulary3D2024a} improves upon OpenMask3D with a 2D-guided 3D-instance proposal module to further optimize initially generated mask proposals. It uses a two-branch system: one generating 3D proposals, as in OpenMask3D, and an additional branch that groups superpoints.
Despite improvements, Open3DIS depends on the pre-trained models Mask3D, Grounding DINO \cite{liuGroundingDINOMarrying2024}, SAM \cite{kirillovSegmentAnything2023a}, ISBNet \cite{ngoISBNet3DPoint2023}, and CLIP. Since all of these are biased towards their respective datasets from pre-training, the use of Open3DIS in out-of-context domains is only feasible when fine-tuning the individual models.
Evidence of effective performance in the industrial domain is largely limited to SAM \cite{moenckIndustrialSegmentAnything2023}.

\subsection{3D scene understanding in industrial environments}

Close to the targeted domain in this work, namely industrial environments, such as manufacturing plants, factories, production systems, workshops, and areas of intralogistics, is 3D scene understanding in process plants:
Multiple works investigate learning-based approaches here, for example, segmenting structural objects of different shapes as well as other plant-related objects, like pipes, valves, or flanges \cite{yinAutomatedSemanticSegmentation2021,agapakiInstanceSegmentationIndustrial2021,imabuchiDiscriminationPlantStructures2024,noichlAutomatedMethodsCreating2025}. The objective is to derive geometric digital twins from laser scans, serving offline retrofit or planning activities. All methods use fully supervised models in the 2D or 3D domain, trained on datasets with a small number of classes. 
Works that directly address the same domain as we do \cite{sommerAutomatedGenerationDigital2023,petschniggPointBasedDeep2020,wangGeoContrastGeometricKnowledgebased2025}, use fully supervised methods trained on either real-world datasets \cite{petschniggPointBasedDeep2020,wangGeoContrastGeometricKnowledgebased2025} or synthetically generated samples from CAD data \cite{sommerAutomatedGenerationDigital2023}.

In the context of open-vocabulary scene understanding in domains other than the popular benchmark datasets, such as ScanNet (which we elaborated on in the previous section), Kamali et al. proposes projecting point clouds to 2D images and using off-the-shelf 2D VLFM for segmentation, followed by back-projecting the positions into the 3D point clouds \cite{kamaliZeroShot3dAsset2025}. However, additional pre-processing steps, like wall removal, are necessary. Further, solutions to viewpoint selection and feature merging strategies are missing.

\section{Preliminary experiments}\label{sec:preliminary}

\begin{figure}[htbp]
    \centering
    \begin{subfigure}{\linewidth}
        \centering
        \includegraphics[width=1.0\linewidth]{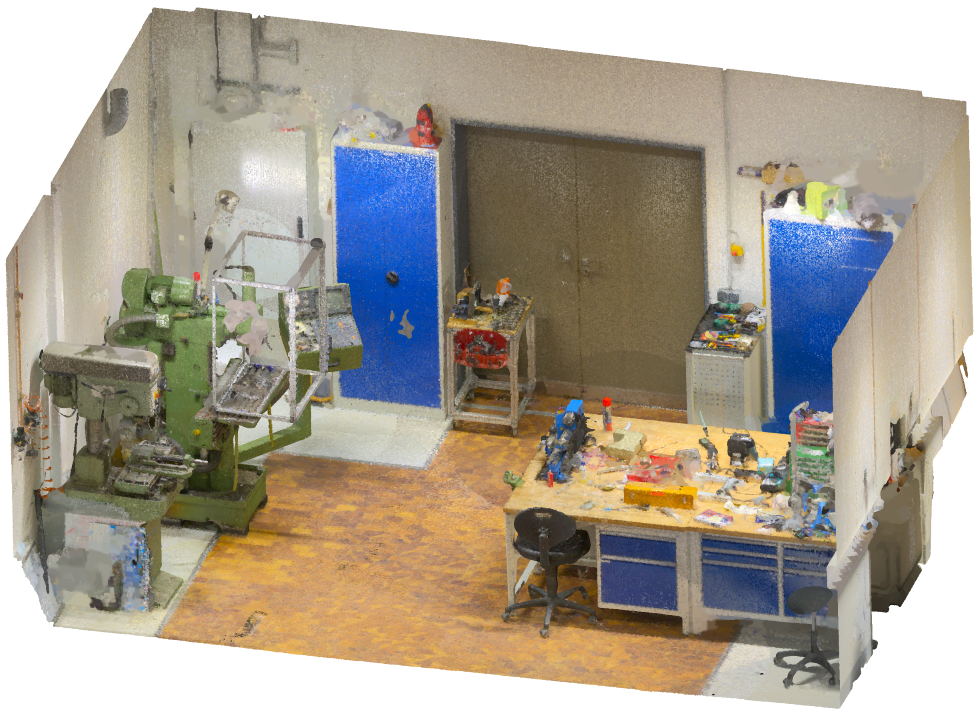}
        \caption{3D scan cut-out}
        \label{fig:workshop_3d}
    \end{subfigure}
    \\[10pt]
    \begin{subfigure}{0.48\linewidth}
        \centering
        \includegraphics[width=\linewidth]{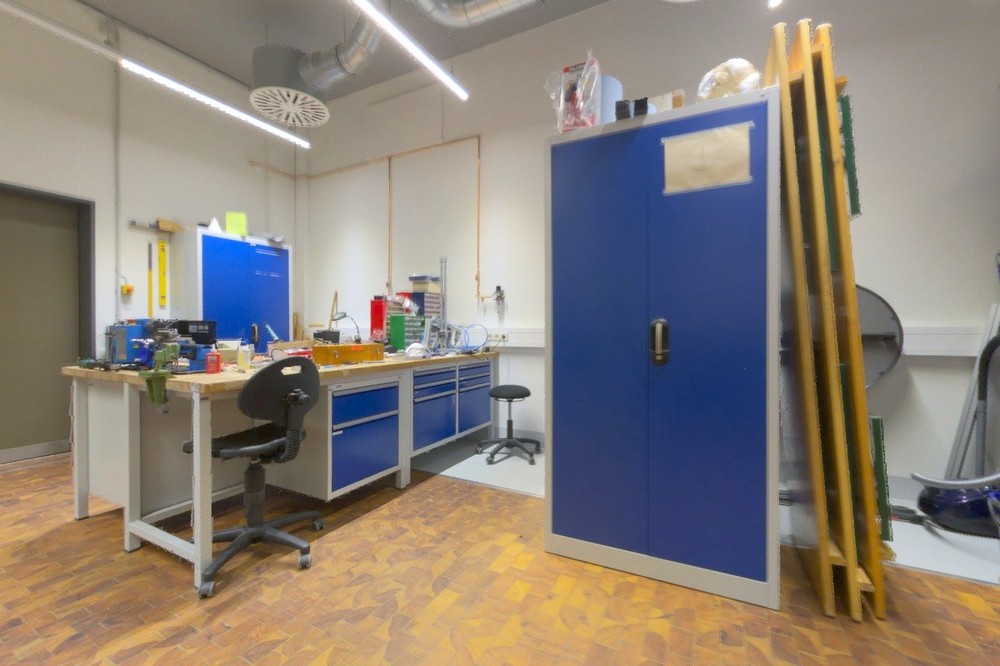}
        \caption{Viewpoint 1}
        \label{fig:workshop_view1}
    \end{subfigure}
    \hfill
    \begin{subfigure}{0.48\linewidth}
        \centering
        \includegraphics[width=\linewidth]{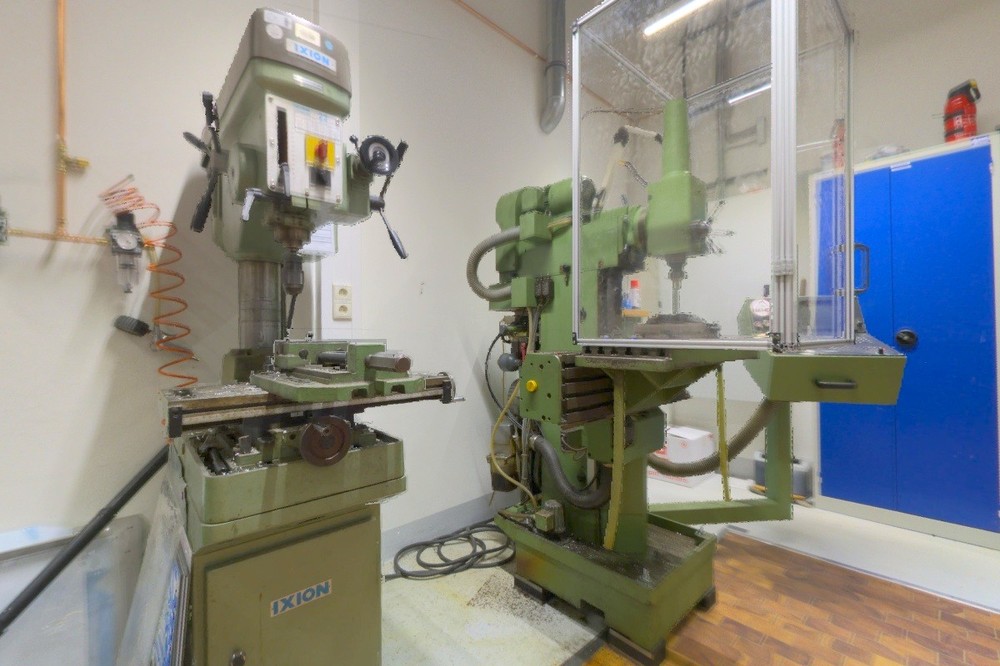}
        \caption{Viewpoint 2}
        \label{fig:workshop_view2}
    \end{subfigure}

    \caption{The industrial scene under study: (a) 3D scan cut-out, and sample images from (b, c) two different viewpoints.}
    \label{fig:workshop}
\end{figure}

\begin{figure}
    \centering
    \includegraphics[width=1.00\linewidth]{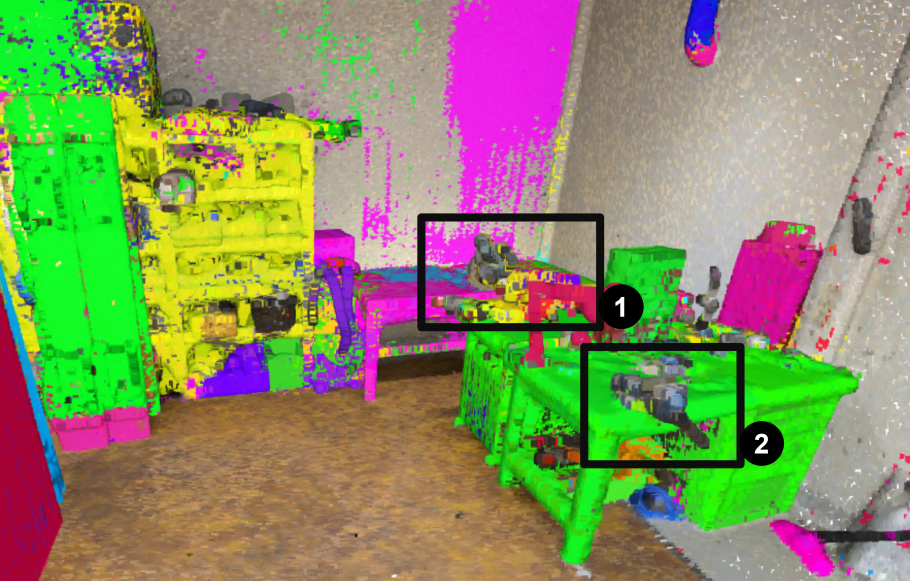}
    \includegraphics[width=1.00\linewidth]{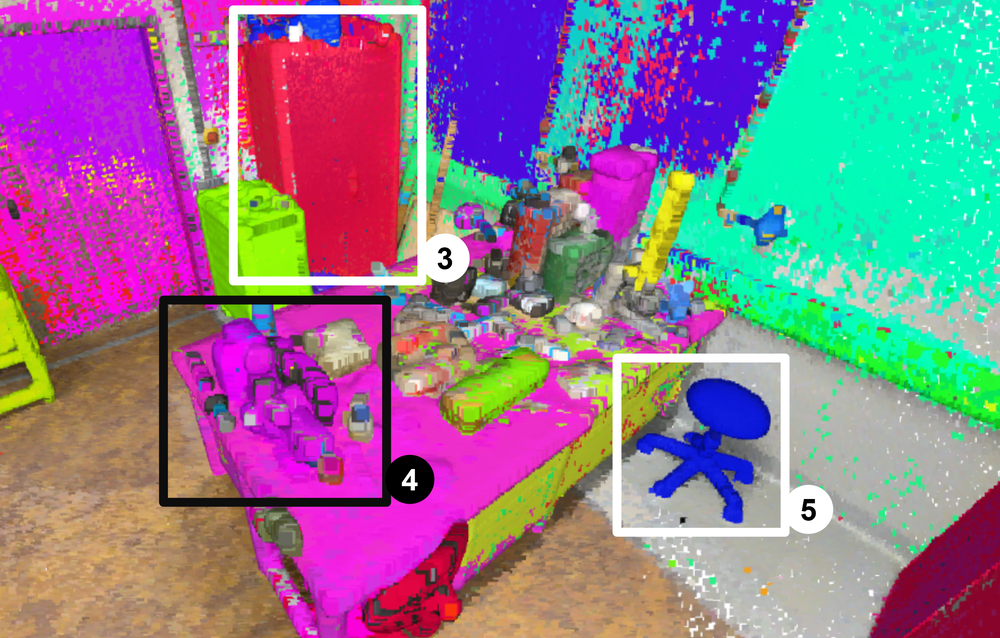}
    \caption{Mask3D-generated mask proposals (color-coded): evaluated as high (white boxes: cabinet [3], chair [5]) and low quality masks (black boxes: table saw [1], vise [2], lathe [4]).}
    \label{fig:workshop_mask_proposals_industrial_scene}
\end{figure}

Experiments in this work are conducted on a 3D scene that consists of a workshop environment as depicted in Fig.~\ref{fig:workshop}, featuring various objects related to the industrial context, such as a lathe, vise, milling machine, as well as smaller hand tools. The 3D scan was captured using a \textit{Leica BLK 360} Terrestrial Laser Scanner (TLS). The scanner captures not only point-wise distance information but also a set of 2D images, which are first concatenated into equirectangular projected panoramic images. From these, we sampled images in $10^{\circ}$ steps, moving along the horizontal axis of each viewpoint.
In total, we acquired $21$ viewpoints.
The point cloud was cleaned and downsampled to a resolution of $5\,\,mm$. Subsequently, we applied surface reconstruction and rendered depth images to allow occlusion checks.

As outlined in Sec.~\ref{sec:related_works_open_vocab}, current open-vocabulary segmentation approaches heavily depend on the class-agnostic instance mask proposals. In multiple works \cite{takmazOpenMask3DOpenvocabulary3D2023b,nguyenOpen3DISOpenVocabulary3D2024a,jungDetailsMatterIndoor2025}, Mask3D is used, pretrained on an indoor household dataset (ScanNet200 \cite{rozenberszkiLanguageGroundedIndoor3D2022}).
In the preliminary experiments, we generated class-agnostic mask proposals in our workshop scene, as depicted in Fig.~\ref{fig:workshop_mask_proposals_industrial_scene}.
The figure clearly demonstrates the significant impact of the dataset context used in pretraining.
The masks' quality strongly depends on the type of object. Objects commonly found in household contexts, such as doors, chairs, or tables, are accurately captured, whereas out-of-context objects, like a lathe, vise, circular saw, or other tools, are not recognized at all.
While correctly detecting household objects (white boxes), Mask3D fails to recognize the ones related to industrial parts (black boxes).

\section{Method}\label{sec:method}
In the following, we present our method, which closely follows OpenMask3D; however, the fundamental difference is that we do not use pre-trained class-agnostic mask/instance proposal networks.
Instead, we use traditionally pre-computed superpoints, which are semantically merged throughout the process.
By doing so, we aim to overcome the limitation discussed in Sec.~\ref{sec:preliminary} regarding out-of-context scenes (e.g., non-household scenes).\\

\begin{figure}
    \centering
    \includegraphics[width=1.0\linewidth]{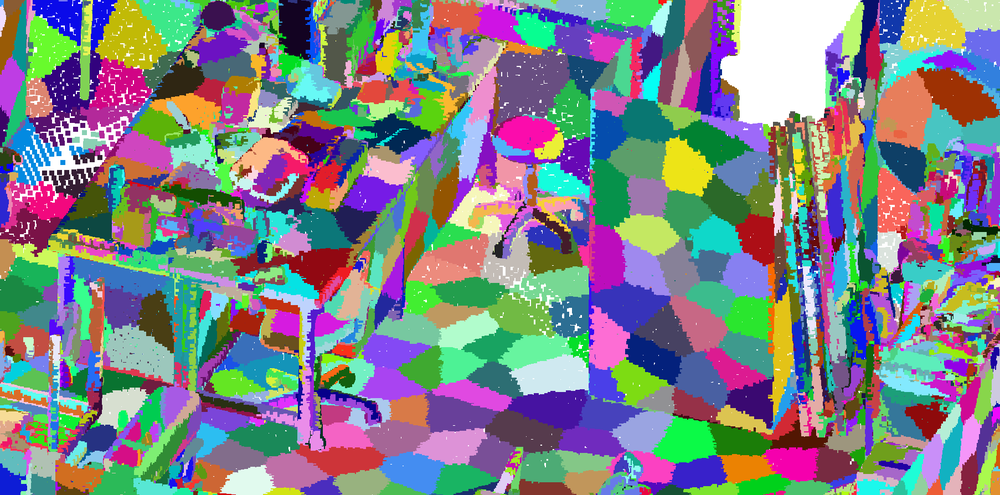}
    \caption{Superpoint-based oversegmentation of the workshop scene.}
    \label{fig:workshop_superpoints}
\end{figure}

\paragraph{Superpoints}
To reduce complexity, a common method is to reduce the points of a point cloud into subsets of superpoints, each treated as a single entity.
We use BPSS \cite{linBetterBoundaryPreserved2018a} for superpoint generation, which formalizes superpoint segmentation as a subset selection problem, seeking to select a specific number of representative points by minimizing an explicit energy function.
Fig.~\ref{fig:workshop_superpoints} depicts generated superpoints in the workshop scene, underlining how the superpoints divide the scene into subsets, while respecting the underlying real-world objects' edges and curvatures.
Afterward, given superpoints per scene, we build an adjacency graph where each superpoint represents a node, while the nodes are connected based on mesh connectivity.

\begin{figure}
     \centering
     \begin{subfigure}[b]{0.48\linewidth}
         \centering
         \includegraphics[width=\linewidth, height=0.9\linewidth]{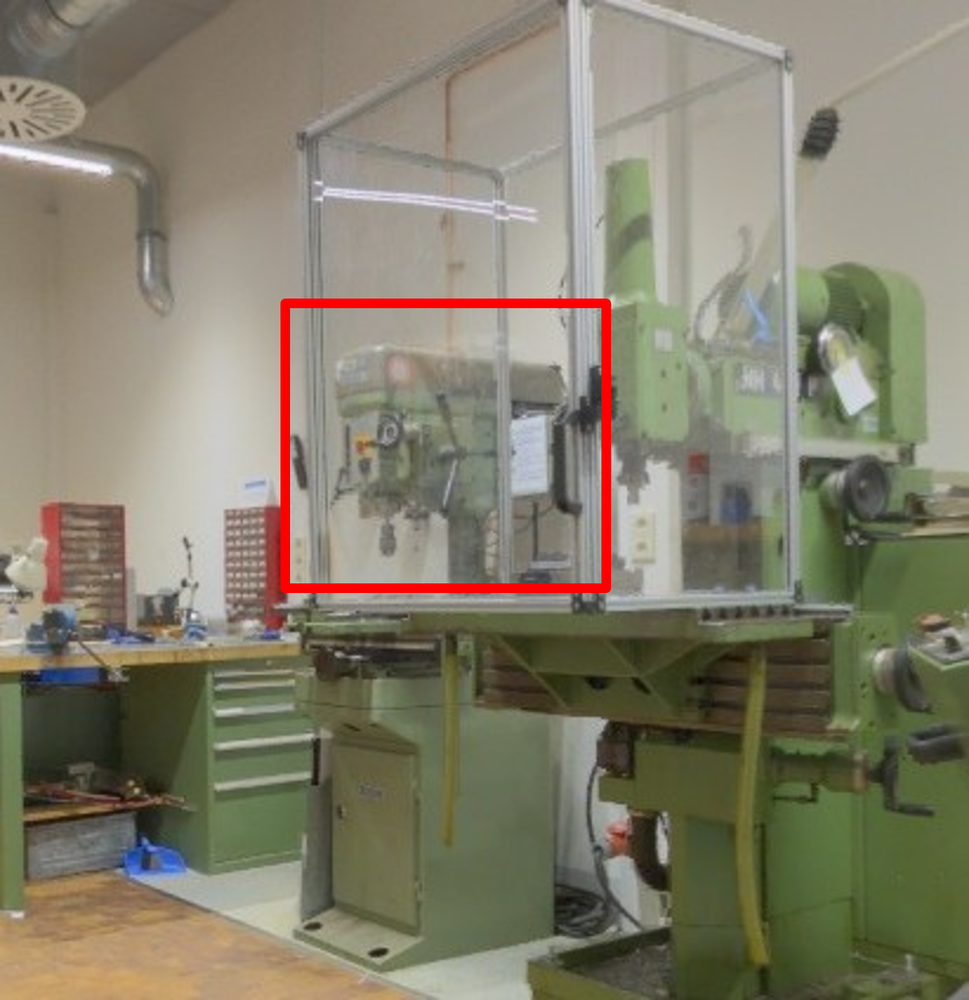}
         \caption{Without masking}
         \label{fig:without_masking}
     \end{subfigure}
     \hfill
     \begin{subfigure}[b]{0.45\linewidth}
         \centering
         \includegraphics[width=\linewidth, height=0.9\linewidth]{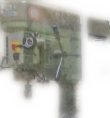}
         \caption{With masking}
         \label{fig:with_masking}
     \end{subfigure}
     
     \caption{Masking leads to a more precise feature representation. Compared to the unmasked version (a), masking (b) makes it clear which specific object is being considered.}
     \label{fig:feature_computation_masking}
\end{figure}

\paragraph{Feature Extraction}
The generated superpoints serve as mask proposals. We project each into all available 2D images, deriving the number of visible points in each frame, while the frames including the highest number of points per superpoint (after occlusion check through depth images) form the top-k views. Next, for each superpoint and its top-k views, we randomly sample a subset of points, which we use as prompts for SAM to generate 2D masks.
We follow the idea from Jung et al., using only the pixels included in the SAM-generated mask, while whitening the rest \cite{jungDetailsMatterIndoor2025}. As exemplified in Fig.~\ref{fig:feature_computation_masking}, masking makes the crops more precise and object-aware, since foreground or background objects are masked out. This is contrary to OpenMask3D; however, since OpenMask3D already uses final instance proposals, the SAM is more effectively guided towards the targeted objects.
Then, we leverage CLIP to extract a feature per superpoint by providing the set of image crops obtained from the previous steps and subsequently averaging the CLIP embeddings.

\paragraph{Superpoint merging}
Based on the superpoints' adjacency graph, we merge neighbors based on their CLIP embedding through cosine similarity.
Superpoint merging results in more coherent masks after multiple iterations, leading to more context-aware CLIP features, which are necessary in the final step, the querying.
We use a value of $\tau=0.95$, which empirically yielded acceptable results for the task of semantic segmentation.
However, it still leads to oversegmentation, requiring clustering of the 3D points in a post-processing step during querying, when targeting instance segmentation.

\paragraph{Open-vocabulary querying}
Finally, after multiple rounds of feature extraction and superpoint merging, we obtain a reduced set of superpoints and their CLIP embeddings. Open-vocabulary querying is then enabled by deriving the cosine similarity of all superpoints' features with the given text embedding.

\section{Experiments}\label{sec:experiments}
We qualitatively conducted experiments on the presented workshop scene. Quantitative evaluation was not possible, as we do not currently possess ground truth data.
In all experiments, we performed $8$ rounds of merging using standard CLIP features, followed by generating features using IndustrialCLIP.
We use IndustrialCLIP, adapted to the industrial domain through prompt learning and fine-tuning an image adapter on the \textit{label short} of the ILID dataset. Further details can be found in the publication \cite{moenckIndustrialLanguageImageDataset2024b}.

\subsection{Single prompt scores}
\begin{figure}
    \centering

    \subfloat["bed without clothes on it" (non-industrial scene)]{
    \includegraphics[width=0.75\linewidth]{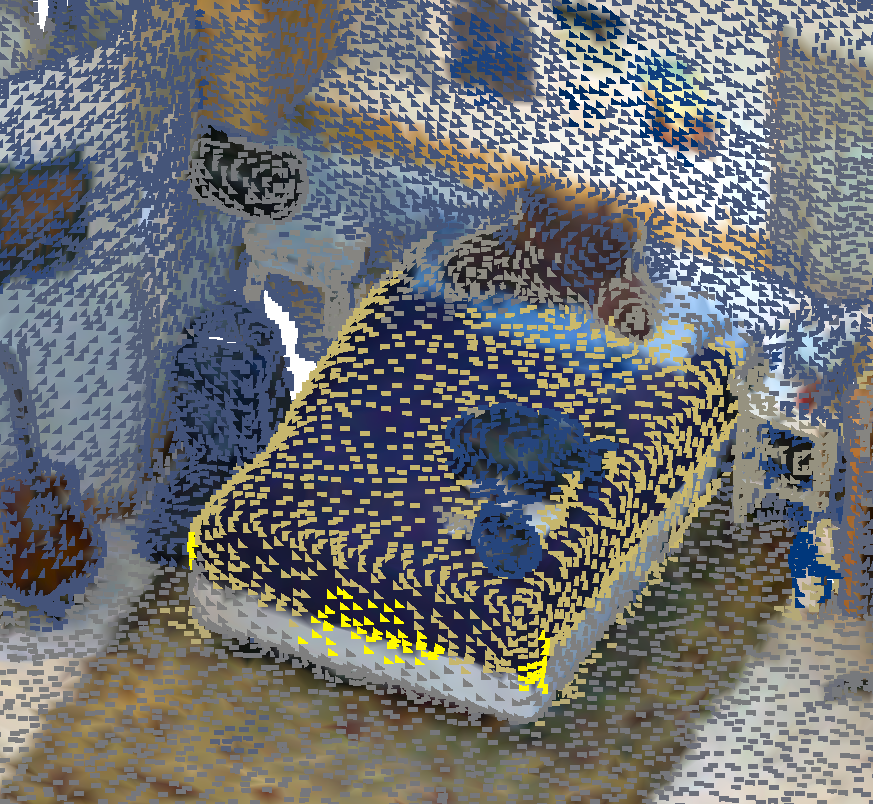}}%
    \\[10pt]
    
    \subfloat["milling machine" (workshop scene)]{
    \includegraphics[width=0.75\linewidth]{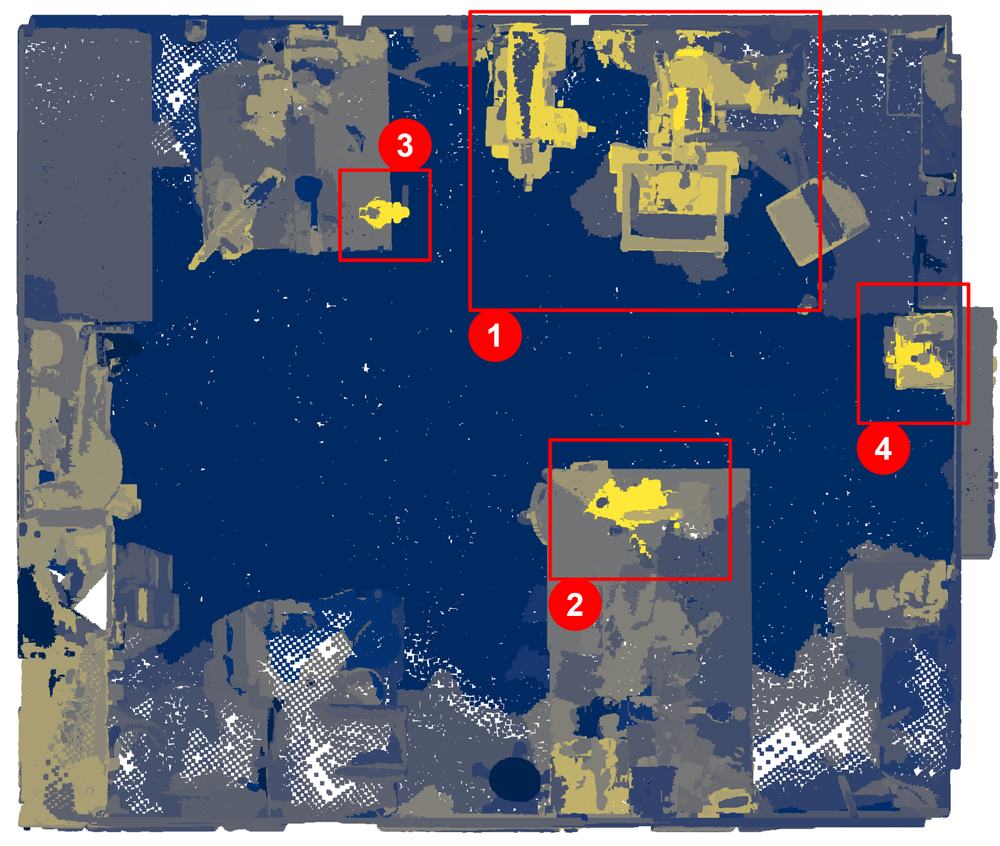}}%

    \caption{Similarity scores (blue and yellow-colored points show low and high similarity, respectively).}
    \label{fig:qualitative_single_prompt}
\end{figure}

Firstly, we discuss prompting in a non-industrial scene (from ScanNet) and then in our industrial workshop scene.
Fig.~\ref{fig:qualitative_single_prompt} depicts two different prompts and the resulting scores per superpoint. Blue and yellow areas indicate low and high cosine similarity, respectively.
Fig.~\ref{fig:qualitative_single_prompt} (a) depicts the result from evaluation on a ScanNet scene that demonstrates the context-aware merging of superpoints, while semantically-distant objects are not unified into one superpoint. Prompting "bed without clothes on it" results in high scores for superpoints lying directly on the bed, while superpoints representing clothes that lie on the bed are not merged with the surrounding superpoints, showing low similarity to the given prompt.

\begin{figure}
    \centering

    \subfloat[Circular saw]{
    \includegraphics[width=0.75\linewidth]{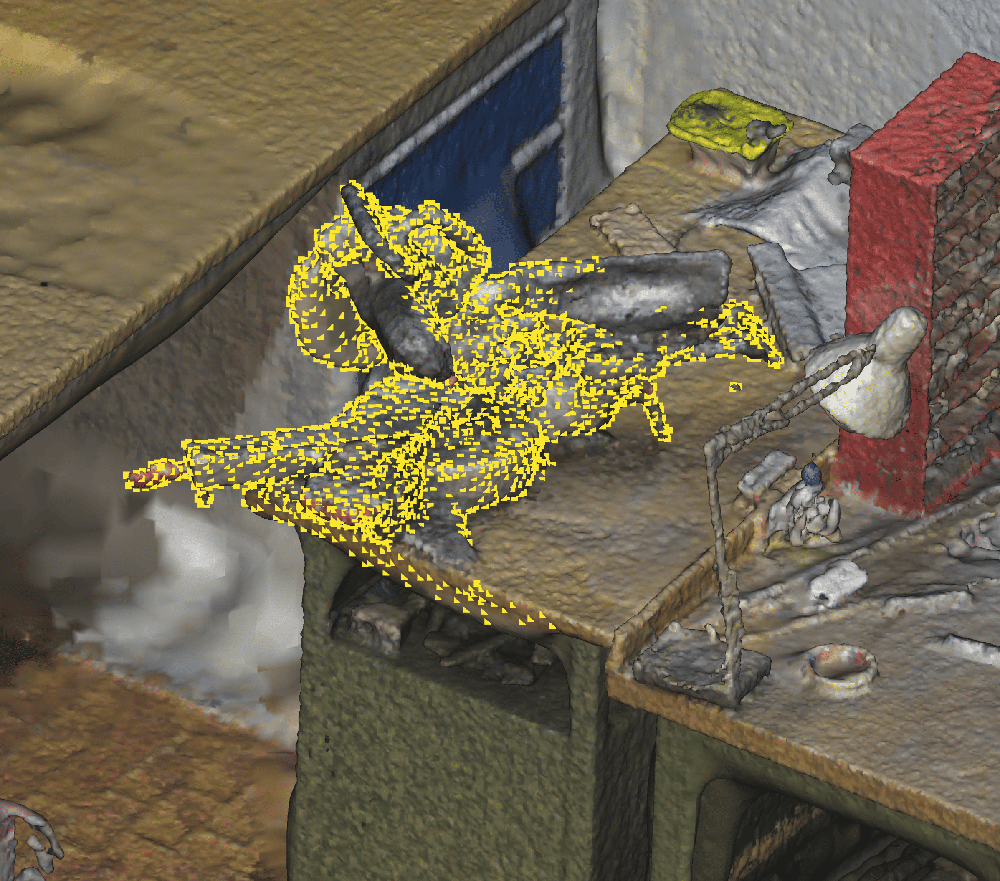}}%
    \\[10pt]
    
    \subfloat[Milling machine]{
    \includegraphics[width=0.75\linewidth]{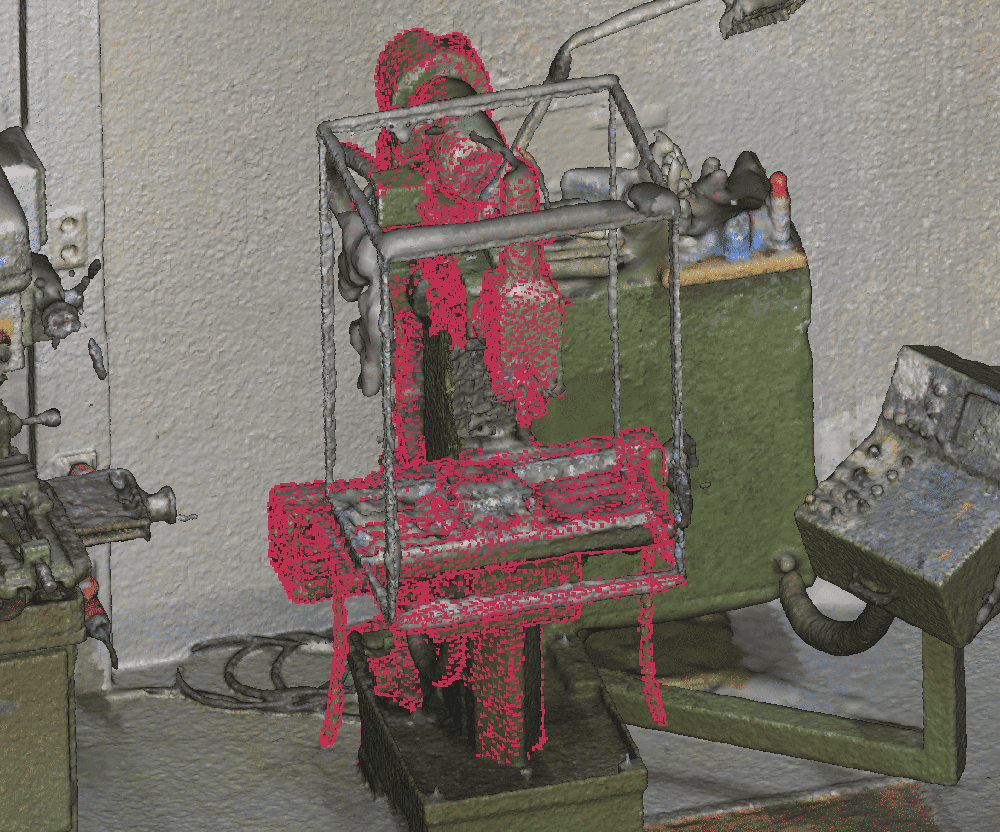}}%

    \caption{Instances after clustering: output masks are highlighted on the reconstructed 3D mesh.}
    \label{fig:qualitative_workshop_results}
\end{figure}

Fig.~\ref{fig:qualitative_single_prompt} (b) depicts the resulting scores when prompting for "milling machine" in the workshop scene. Three objects directly show high scores: the drilling and milling machine (1) and the lathe (2). However, the vise (3) and a set of collets on a trolley show higher scores compared to the rest of the scene. Although the prompt can be seen as unambiguous, based solely on vision information, the difference between a drilling and milling machine is intricate, highlighting the contextual limitation of IndustrialCLIP.
The other objects with high scores (lathe, vise, and collets) are semantically close to the prompt, but can be counted as false positives. Despite that, no "milling machines" are missed (false negatives). This highlights IndustrialCLIP's limitations in terms of language breadth and contextual information.
Given that the model has not been explicitly trained on the prompted classes, the results highlight the potential of VLFM for 3D scene perception in an open-vocabulary setting.
Finally, we used HDBSCAN to cluster the thresholded segmentation results to obtain instances, as shown in Fig.~\ref{fig:qualitative_workshop_results}.

\subsection{Comparison to CLIP}
We further examined the difference between CLIP and IndustrialCLIP features. Fig.~\ref{fig:qualitative_clip_iclip} depicts the same area of the workshop scene and the scores per point after prompting "vise". IndustrialCLIP shows high scores, while CLIP features are merely different between the actual object and its surroundings.
While this is a positive example, in general, we find that IndustrialCLIP achieves better results for objects related to the industry.
This, in turn, leads to an overall lower quality of masks for every other non-industrial object.
That is also the reason why using IndustrialCLIP features during the superpoint merging process failed, and we omitted using it for further experiments in merging.
Generally speaking, IndustrialCLIP overfits to industrial objects, which are in a similar setting as displayed in catalog images.

\section{Discussion and conclusion}\label{sec:discussion}

In this work, we presented and evaluated a training-free, open-vocabulary 3D perception approach in the context of industrial environments.
We demonstrated that existing approaches, which typically rely on class-agnostic instance segmentation models pre-trained on household scenes, fail to generalize to out-of-context industrial settings.
To overcome this limitation, we proposed a method that generates mask proposals by iteratively merging superpoints based on their semantic features. This approach avoids reliance on pre-trained 3D models and enables context-aware segmentation, as demonstrated in our qualitative experiments.

Furthermore, we evaluated the effectiveness of IndustrialCLIP, a Vision-Language Foundation Model (VLFM) adapted for the industrial domain. Our findings indicate that while IndustrialCLIP provides superior feature representation for specific industrial objects (like a "vise") compared to standard CLIP, it also exhibits limitations. These include difficulty distinguishing between semantically similar objects (e.g., milling machine vs. drilling machine) and a tendency to overfit to industrial catalog-style imagery.

The results highlight the potential of domain-adapted VLFM for 3D perception but also underscore the challenges of overfitting and maintaining broad contextual understanding.
More complex domains require support for more language-broader prompts.

\section*{CRediT author statement}
\noindent
Conceptualization \& Methodology: KM, AF; Software \& Data: AF, KM; Supervision \& Administration: KM, TS; Funding: TS; Visualization: KM, AF; Writing – Original Draft: KM, AF; Writing – Review \& Editing: All authors.

{
    \small
    \bibliographystyle{elsarticle-num}
    \bibliography{main}
}

\end{document}